\documentclass[conference]{IEEEtran}
\IEEEoverridecommandlockouts
\usepackage{cite}
\usepackage{amsmath,amssymb,amsfonts}
\usepackage{algorithmic}
\usepackage{graphicx}
\usepackage{textcomp}
\usepackage{xcolor}
\usepackage{fancyhdr}

\def\BibTeX{{\rm B\kern-.05em{\sc i\kern-.025em b}\kern-.08em
    T\kern-.1667em\lower.7ex\hbox{E}\kern-.125emX}}

\begin{document}

\title{
Knowledge Graph Enhanced Event Extraction in Financial Documents\\
}
\author{
\IEEEauthorblockN{Kaihao Guo,  Tianpei Jiang, Haipeng Zhang\IEEEauthorrefmark{1}\thanks{\IEEEauthorrefmark{1}Haipeng Zhang is the corresponding author.}} 
\IEEEauthorblockA{\textit{School of Information Science and Technology, ShanghaiTech University} \\
Shanghai, China \\
\{guokh, jiangtp, zhanghp\}@shanghaitech.edu.cn
}

}

%\author{\IEEEauthorblockN{Kaihao Guo}
%\IEEEauthorblockA{\textit{School of Information Science and Technology,} \\
%\textit{ShanghaiTech University} \\
%Shanghai, China \\
%guokh@shanghaitech.edu,cn}
%\and
%\IEEEauthorblockN{Tianpei Jiang}
%\IEEEauthorblockA{\textit{School of Information Science and Technology,} \\
%\textit{ShanghaiTech University} \\
%Shanghai, China \\
%jiangtp@shanghaitech.edu,cn}
%\and
%\IEEEauthorblockN{Haipeng Zhang\IEEEauthorrefmark{1}}
%\IEEEauthorblockA{\textit{School of Information Science and Technology,} \\
%\textit{ShanghaiTech University} \\
%Shanghai, China \\
%zhanghp@shanghaitech.edu,cn}
%}
\maketitle
\thispagestyle{fancy}
\fancyhead{}
\lhead{}
\cfoot{}
\rfoot{}

\begin{abstract}
%本文提出了一种基于知识增强的事件抽取模型，我们使用了知识图谱作为先验知识库，并在中文金融公告事件抽取任务上达到了state of art的结果
Event extraction is a classic task in natural language processing with wide use in handling large amount of yet rapidly growing financial, legal, medical, and government documents which often contain multiple events with their elements scattered and mixed across the documents, making the problem much more difficult. 
Though the underlying relations between event elements to be extracted provide helpful contextual information, they are somehow overlooked in prior studies. 

We showcase the enhancement to this task brought by utilizing the knowledge graph that captures entity relations and their attributes. 
We propose a first event extraction framework that embeds a knowledge graph through a Graph Neural Network and integrates the embedding with regular features, all at document-level.
Specifically, for extracting events from Chinese financial announcements, our method outperforms the state-of-the-art method by 5.3\% in F1-score.

\end{abstract}

\section{Introduction}
%介绍本文工作。本文动机为非结构化的金融公告文本数量越来越多，而其中蕴含的事件信息是文本描述的主要内容。一个自动有效的从非结构化文本抽取结构化信息的系统可以节省时间和人力资源。同时，抽取出的结构化信息可以帮助我们更好的分析数据以及监控市场风险的形成。
%为什么要使用知识图谱：知识图谱（Knowledge Graph）是一种有效的知识表示方式。相比传统知识库，知识图谱以图的形式对知识进行存储，从而可以处理高纬度的查询内容，并且可以利用图的特征来处理数据。知识增强（knowledge enhancement）是指使用先验知识来帮助完成任务的方法。知识增强目前已经运用在了很多NLP任务上，包括命名实体识别，情感分析以及长文本生成任务。

%本文的contribution有两种：提出了一个先验的事件抽取模型，提供了一个远程监督方式标注的中文金融公告诉讼类型事件抽取数据集
%我们实现了一个端到端的事件抽取模型（Event Extraction）。事件抽取的模型使用了NER+知识图谱知识增强+事件路径（名字还没想好，暂定KGDee）构建三部分内容。
%Graph如何引入：我们使用了两种的方式来引入graph信息：对关系进行one-hot embedding以及利用GNN进行node embedding。我们对节点embedding如何加入模型中也使用了不同的方式，包括了使用single linear layer，attention layer两种方式。
%我们同时还利用了基于知识图谱的远程监督来进行数据标注。我们通过对公告进行模板匹配来抽取出一个事件图谱，并用事件图谱在大范围的新闻以及公告中抽取出了 2,937条诉讼类型事件训练集
%数据描述部分：我们的知识图谱的数据是从天眼查上获得的，包括了3000家上市公司的数据。其中有347570个实体节点以及441127条关系。（关系类型有7种，包括分支结构，授信，拥有者（holder），投资（investor），法人（LegalPerson），质押（Pledge）以及职工（Staff））。

%介绍知识图谱以及知识增强
Event extraction is a classic yet challenging task, which obtains key elements with attributes and relations between elements from documents.
Take the equity pledge event in financial announcements for example, it contains various elements that are scattered across the document, including pledged shares, percentage of shares, start and end time, pledgers, and pledgees. 
Besides, there are relations between entities that can be crucial and yet error-prone. %此外，实体之间存在一些至关重要的关系，而且容易出错。
For instance, the pledger-pledgee relation is easily mistaken because the two words pledger and pledgee have multiple aliases that may cause confusion in Chinese.
Furthermore, it is common that a document contains multiple events which makes the task much more complicated – one event’s elements can be confused with the elements of other events. 

Intuitively, external information can help event extraction tasks. Knowledge bases, such as the FrameNet lexical database, are utilized to label events~\cite{liu2017exploiting}. In a more specific domain, tree-structured long short-term memory networks have been used to capture gene tree dependency information in order to extract biomedical events~\cite{li2019biomedical}. However, the rich relational and structural information from knowledge graphs that links elements has not been fully exploited for enhancement in this scenario.

On the basis of a state-of-the-art document-level event extraction method~\cite{zheng2019doc2edag}, we propose a framework that integrates a prior knowledge graph to extract events from long documents. It consists of three steps. First, we use a graph encoder to embed the knowledge graph information for entities in our graph. Second, we perform named entity recognition (NER) to identify possibly relevant entities. Finally, we use two different approaches, Transformer layer~\cite{vaswani2017attention} and linear layer, to input the graph embedding of the extracted entity to the event extraction model to help the tasks.

We apply the framework to perform event extraction on the Chinese financial announcements of the listed companies in China, in which we identify and extract six types of events (equity freeze, equity repurchase, equity overweight, equity underweight, equity pledge and lawsuit) that may indicate risks and require immediate actions from market regulators and financial institutions that have corresponding companies in their portfolios. 
Our dataset consists of 21,871 financial announcements from over 3,000 listed companies. With regard to the knowledge graph, we collect directional relation data for these companies, covering 347,570 entities (companies and persons) and 441,127 relations (e.g. share-holding relation, managing relation) which are publicly available. 

Zheng \emph{et al.}\cite{zheng2019doc2edag} extract 5 types of the 6 aforementioned events and provide corresponding dataset labeled with event details. The lawsuit events, which they have not covered are often accompanied by risks. In order to better apply the model to the real-world scenario, we use the distant supervision (DS) method based on a knowledge graph to label the announcement data and construct a lawsuit type event dataset.

The main contributions in this paper are two-folded:

1.	We propose a framework that uses prior knowledge from basic knowledge graph to assist automated information extraction that outperforms state-of-the-art baselines;

2.	We construct and release a new lawsuit event dataset for event extraction\footnote{http://bitly.ws/auMb}.
%\footnote{\url{http://bitly.ws/auMb}}.
	
The paper is organized as follows.
Section \ref{related work} shows the related work in event extraction, knowledge graph and graph neural network.
Section \ref{problem definitions} shows our definition of the problem, including the definition of the knowledge graph embedding, document-level event extraction task, and the description of the predefined event roles and types.
Section \ref{methodology} shows the main algorithms in our framework, including the graph embedding method and the encoder, event extraction and how to use distant supervision to label the data.
Section \ref{experiments} describes our data, the experiment setup, and our results compared with a state-of-the-art method. Finally, we conclude our work in Section \ref{conclusion and future work}.
%In event extraction (EE) tasks, events are defined with event types, event arguments, and argument role data structures. The purpose of EE is to extract structured event information from unstructured text. In recent years, with the development of the internet, the amount of unstructured information data continues to increase. So that, EE become increasingly important. In China’s financial market, listed companies play very important roles. Any changes in listed companies may affect a large number of companies. The announcements of listed companies are an important source of information for analyzing and supervising listed companies. Since 2007, the number of listed company announcements increasing every year, and about 400,000 listed company announcements are released in 2019. Such a large number makes it difficult to analyze manually and extract information from the text. Event extraction can help us automatically extract events from massive amounts of data, help people avoid risks or find valuable investment opportunities.

\section{Related Work}\label{related work}
%引入部分
Our work is inspired by three threads of research: event extraction, knowledge graph and graph neural networks.
\subsection{Event Extraction}
%历史（ACE2005数据集，句子级别事件抽取，篇章级别事件抽取）
There has been a lot of studies on event extraction (EE), but the previous research focuses on sentence-level extraction (SEE) with most experiments performed on the ACE2005 dataset\footnote{https://www.ldc.upenn.edu/collaborations/past-projects/ace}, which contains labeled events in English, Arabic and Chinese. As a first systematic SEE framework, Ahn~\cite{ahn2006the} divides EE tasks into multiple sub-tasks, and proved that a series simple machine learning approaches can achieve the performance of best systems. 
Liao and Grishman~\cite{liao2010using} use document-level information for EE tasks, but their method still cannot handle the scenario where the event arguments are scattered in different sentences. Hong \emph{et al.}~\cite{hong2011using} propose a method using cross-entity inference, but the authors only consider the entity types information and ignore the other information of entities. With the development of deep learning, Chen \emph{et al.}~\cite{chen2015event} propose a multi-pooled convolution neural network EE method, which become a standard method for SEE tasks. However, these aforementioned methods ignore the external information that may be useful in the tasks.

%修改句式， 承接关系
%\footnote{\url{https://www.ldc.upenn.edu/collaborations/past-projects/ace}}
%However, SEE that utilizes only information within a sentence cannot solve the problem of argument dispersal in our scenario where information associated with the given event is usually scattered across an article and one article often contains multiple events.% 并到前面一段，原因先提

Document-level event extraction (DEE) is the method of extracting elements from the full text and combining the elements into events. Compared with SEE, DEE is designed to solve the problem of argument dispersal. At present, there is little research on DEE. One of the major difficulties is associated with the lack of enough document-level labeled datasets. The DCFEE model~\cite{yang2018dcfee} uses a distant supervised method to label data, and utilizes the sequence labeling model combined with the key sentence extraction model to extract the events. The latest Doc2EDAG model~\cite{zheng2019doc2edag} performs named entity recognition (NER) and encodes the entity and the sentence which contains the entity. An entity’s path generation method, which is a Transformer, is then used to extract events. Though these DEE methods mitigate the argument dispersal problem, the external knowledge which indicates the relations between entities is somehow also overlooked. 
%金融文本

\subsection{Knowledge Graph Enhanced Information Extraction}
%Knowledge graph is a kind of knowledge base stored in the form of triples~\cite{bizer2009dbpedia}. The relations between entities are often represented by knowledge graphs. 
Knowledge graph has been a powerful tool for representing the knowledge by entities and relations between them, which is usually stored in the form of triples~\cite{bizer2009dbpedia}. A knowledge graph with rich information can enhance the performance of the task in event extraction. Ruan \emph{et al.}~\cite{ruan2016building} incorporate cross-genre knowledge to improve the information extractor performance for social media by 15\% in F-score. Li and Ji~\cite{li2016cross} propose a distance supervised relation extraction method, where a knowledge graph with various semantic correlations among entities is embedded to help extracting information from long-tailed and imbalanced data. Zhang \emph{et al.}~\cite{zhang2019long} develop a system of knowledge-driven information extraction in Russian with a set of linguistic resources introduced. These works have proved the effectiveness of external knowledge in helping information extraction, but their potential improvements in event extraction, which is a quite different task, are barely touched.

%what they do is quite different tasks with ours. %quite different tasks, 有什么关系，找到共同和不同的点，说他们的用处

%In our task, we use a knowledge graph, which consisted of listed companies, people, organizations, industries, and the relation between entities in the graph, to . 
%介绍knowledge graph
%DBpedia~\cite{bizer2009dbpedia}, Yago~\cite{suchanek2007yago}, and WikiData~\cite{2014wikidata}. The industry knowledge graph focuses on a specific industry and is used to assist in analyzing the industry. In~\cite{pujara2017extracting}, the authors use open information extraction technology to extract triples from financial statements for the construction of financial knowledge graphs to predict the relevance of each potential relation. The generation method of knowledge graph also includes automatic machine extraction. NELL~\cite{mitchell2015never} uses machine learning algorithms  to continuously generate new extraction templates and extract structured knowledge from unstructured web page text to form knowledge graph.

\subsection{Graph Convolution Network}
%使用GNN进行Embedding
Graph Convolution Network (GCN) is a common method to embed the knowledge stored in graph structure. The recent advance in the GCN can be categorized as spectral and non-spectral approaches.
%In this section, we mainly mention the Graph Convolution Network (GCN) and related graph embedding method.

%GCN将CNN引入了Graph domain中。从而可以处理以图谱形式存储的Non Euclidean Structure data。目前实现GCN的方式大致分为spectral approaches和non—spectral approaches.
%Spectral approaches使用一种信号变换的方式处理图结构数据。研究b15提出了将图片作为信号，利用拉普拉斯算子来对图进行特征分解的方法。 研究b16 

Spectral approaches use a spectral transformation to deal with graph structure data. Bruna \emph{et al.}~\cite{bruna2014spectral} propose a construction based on the spectrum of the graph Laplacian~\cite{belkin2001laplacian} to learn convolution layers on low-dimension graphs. Kipf \emph{et al.}~\cite{kipf2017semi}, design a model which only operates the filter on the target node and its neighbor nodes to learn a hidden layer representation that encodes the local graph structure and node features. The disadvantage of spectral approaches is that they depend on the structure of the graph. As a result, a model trained on a specific structure cannot be directly applied to graphs with different structures.

Non-spectral approaches of GCN perform convolution by only considering close neighbor nodes of the target node. For instance, Duvenaud \emph{et al.}~\cite{duvenaud2015convolutional} use a series of weight matrices to allow end-to-end learning on graphs, while Hamilton \emph{et al.}~\cite{hamilton2017inductive} propose an inductive framework that learns a function to generate embeddings by aggregating features. These non-spectral approaches lack flexibility when the numbers of neighbors vary and a general operator is desirable. In order to solve this problem, 
Veli\v{c}kovi\'c \emph{et al.}~\cite{2018graph}, implement the graph attention networks~(GATs) model, which introduces an attention mechanism to extract the features of all neighbor nodes. The GATs model only depends on the edge instead of the complete graph structure and can be flexibly applied on directed and undirected graph.

\section{problem definition}\label{problem definitions}

\begin{table*}[tbp]
\caption{Event Types and Their Corresponding Event Roles.}
\begin{center}
\begin{tabular}{|c|c|}
\hline
Event Type& 
Event Roles\\
\hline
Equity Freeze & 
Equity Holder, Froze Shares, Legal Institution, Total Holding Shares, Total Holding Ratio, Date, Unfroze Date \\

Equity Repurchase& 
Company Name, Highest Trading Price, Lowest Trading Price, Repurchased Shares, Closing Date, Repurchase Amount \\

Equity Overweight& 
Equity Holder, Trading Shares, Date, Later Holding Shares, Average Price \\

Equity Underweight& 
Equity Holder, Trading Shares, Date, Later Holding Shares, Average Price \\

Equity Pledge & 
Pledger, Pledged Shares, Pledgee, Total Holding Shares, Total Holding Ratio, Total Pledged Shares, Date \\

Lawsuit & 
Plaintiff, Defendant, Legal Institution, Date \\
\hline
\end{tabular}
\label{eventrole}
\end{center}
\end{table*}

%我们的任务是从中文公告文本中抽取出结构化的事件信息。我们有一个由6种不同类型事件组成的数据集，其中每种事件都有一个预定义的event role表。我们预先判断文本中能抽取出的事件类型，并且将抽取出的实体填充入对应事件类型的event。 role表中。此外，我们还构建了一个与抽取内容相关的上市公司知识图谱。为了利用知识图谱的信息辅助事件抽取任务，我们使用独热编码以及图神经网络两种方法对图中的信息进行嵌入。
%Our task is to extract structured event information from the Chinese announcement text. We have a dataset consisting of six different event types, each of which has a predefined event role table. We predefine the event type belongs to the document, and fill the extracted entities into the event role table of the corresponding event type. In addition, we constructe a knowledge graph of listed companies related to the extracted document. In order to use the information of the knowledge graph to assist the event extraction task, we use one-hot encoding and graph neural network method to embed the information in the graph.

Our task is to extract structured event information from the financial announcement text in Chinese by a knowledge enhancement framework.
In event extraction (EE) tasks, events are defined as structured data event types, event arguments, and event roles. 
Event role is a field which is predefined for each event type. In addition, we define the key roles for each event type which are indispensable in the corresponding event type. Table~\ref{event example} shows an equity pledge event example.
Event argument is the entity which plays an event role in a specific event.
We obtain a dataset consisting of six different event types. 
For each type, there is a list of predefined event roles as shown in Table~\ref{eventrole}. When dealing with an announcement, we determine the event type of the document and then fill the extracted entities into the event role list of the corresponding event type. To be specific, we use the dataset from the Doc2EDAG paper~\cite{zheng2019doc2edag} and we further complement it with the lawsuit event data which is obtained by distant supervision labelling to be described in Section~\ref{methodology}.

\begin{table}[tbp]
\caption{An Equity Pledge event example}
\begin{center}
\begin{tabular}{|c|c|}
\hline
Event Roles& 
Event Argument\\
\hline
Pledger* & 
Zhu Mingqiu \\

PledgedShares*& 
4,000,000 shares \\

Pledgee*& 
Huatai Securities Asset Management 
Co., Ltd. \\

TotalHoldingShares& 
131,573,433 shares \\

TotalHoldingRatio & 
8.47\% \\

TotalPledgedShares & 
55,265,780 shares \\
StartDate & 
2017-07-19 \\
EndDate & 
Null \\

ReleasedDate & 
Null \\
\hline
\end{tabular}
\label{event example}
\end{center}
\end{table}

To evaluate the performance of the event extraction algorithms, we use the micro precision, recall and F1-score as evaluation metrics. To be specific, we compare the extracted event with the ground-truth event which shares the largest overlap in terms of number of event arguments. For each event role, we calculate the true positive rate, false positive rate and false negative rate. Besides, we further evaluate the model on multi-event documents and single-event documents, respectively.

\section{Methodology}\label{methodology}
%在这一部分，我们首先介绍了我们使用DS从公告数据中抽取诉讼类型事件集的流程。%然后我们介绍了我们如何从知识图谱中获得实体的embedding。我们使用了one-hot encoder和GAT两种不同的方法。最后我们介绍了事件抽取模型以及如何在模型中使用先前学习到的graph embedding.
The workflow of the framework is described in this section. We first show the process of employing the distant supervision (DS) method to extract lawsuit events from the announcements. We then describe two different methods to learn the embeddings of entities from the main knowledge graph: one-hot encoder and graph attention network. Finally, we demonstrate the event extraction model and the way of using the previously learned graph embeddings in the model.

Recall that one aim of the framework is to incorporate external knowledge to enhance the performance of the event extractor. Our main knowledge graph stores the external knowledge including key information of China’s listed A-share companies. 

\begin{figure*}[htbp]
\centerline{\includegraphics[width=7 in]{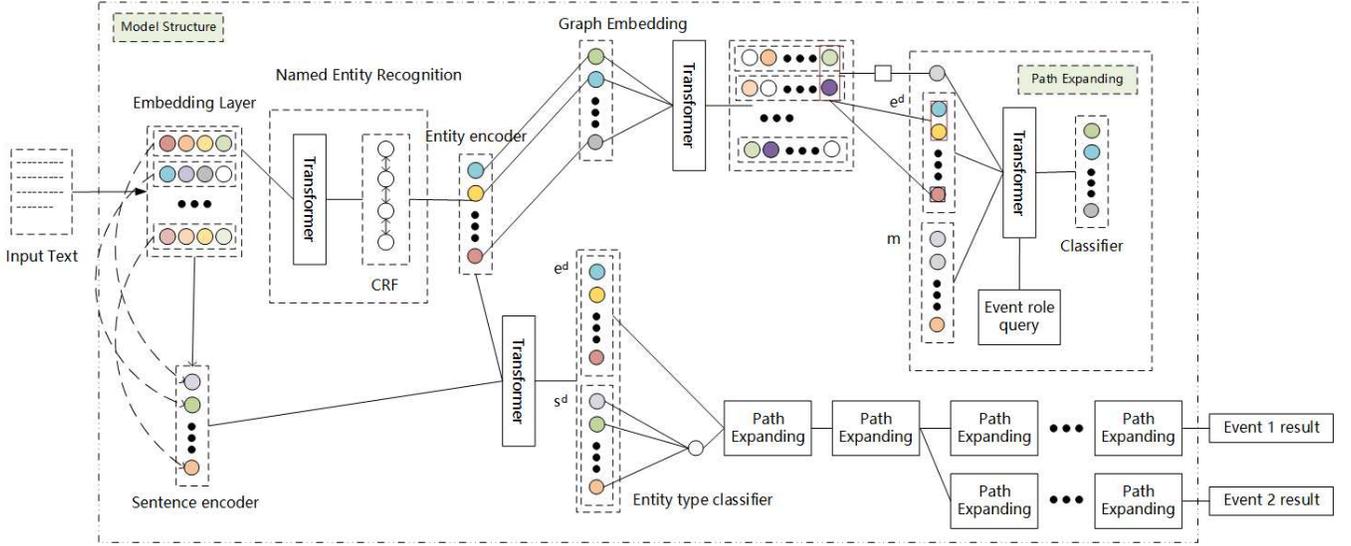}}
\caption{The Transformer encoder approach of the event extraction model. The input of our model is a piece of text, and the output is the event results. In this example, two branches are generated on the second path-expanding task, and finally two different events are extracted.}
\label{ee model}
\end{figure*}

As illustrated in Figure~\ref{ee model}, the framework consists of four major steps:

(1) We embed the entities in the main knowledge graph by their relation and then get an vector that focuses on describing the relations between entities.

(2) We perform named entity recognition (NER) to get the list of candidate entities.

(3) We correspond the extracted entities to the entities on the knowledge graph to obtain the embeddings of these entities. Then we use an encoder to process the embedding of these entities.

(4) We embed the entities and sentences in the article, and use a series of path-expanding tasks to extract events. In path expanding tasks, we use the embeddings obtained from the knowledge graph.

\subsection{Unstructured text labeling based on knowledge graph}\label{ds labelling method}
%我们使用了基于远程监督的算法来抽取诉讼类型事件,步骤：1使用模板匹配的方式从文章中匹配出结构化事件数据。2 通过数据构建事件图谱。3使用远程监督方式来抽取训练数据。

%%换一些比较general的说法，把specific的地方放在experiment里面

The task of document-level event extraction requires labelled data. To increase the various of the dataset, we use a knowledge graph based Distance Supervision (DS) method to label new type of data~\cite{smirnova2018distant}.  

It is worth mentioning that in this part, we use a new DS knowledge graph constructed from data matched by the template. The relations in the DS knowledge graph and the main knowledge graph used in the event extraction are exclusive.

Here we describe the three steps to labeling dataset based on the DS knowledge graph. 
(1) The first step is template matching, which is to ensure high accuracy on the event extraction tasks. We create some templates for events of the selected type. Through these templates, we can extract events from the structured text in the corresponding announcement.
(2) The second step is to construct a DS knowledge graph. The extracted information in the first step is transformed into triples to describe the entities, relations, and attributes of events. Disambiguation is a main problem in constructing the DS knowledge graph. That is, a same entity can have different names in different announcements or even in a same one. We employ a series of rules, such as removing location names (e.g. Shanghai) and suffixes of companies or courts (e.g. Co., Ltd.) to create a new expression of the name. We also crawl a list of company abbreviations to resolve the entity disambiguation problem.
(3) In the third step, we use a DS method based on the DS knowledge graph to label new type of data. For each event type, the parameters of the event is defined as:
\begin{equation} \mbox{Event}_p = \{a_{p1} … a_{pi}, b_{p1}…, b_{pj}\} \label{eq}\end{equation}
where $a$ and $b$ represent key and non-key roles of the event type respectively, $p$ is the event type, and $i$ and $j$’s are indices of event arguments.
We set two constraints in this step. First, we check whether all the predefined key roles of the event appear in the text. Second, we only keep the results whose ratios between extracted arguments and the amount of all possible event roles are above a certain threshold. 

\subsection{Knowledge graph embedding}
%我们使用了两种不同的方法进行知识图谱先验信息的嵌入处理。第一种是简单的独热编码，第二种是图神经网络。
We use two different approaches to embed the information of the main knowledge graph. The first is one-hot encoder based on the connecting paths between two entity nodes and the second is a graph neural network.

\subsubsection{One-hot Encoder}
%我们使用了独热编码来对两个节点之间的关系进行编码。对每两个节点，我们从所有节点a到节点b的路径中选取最短的三条。
%对于每条路径，我们从中提取出所有的关系，并将其以独热形式编码为向量。然后，我们将
For each pair of two target nodes, we select the first $k$ shortest paths $\{P_1,P_2,...,P_k\}$. $P_i = \left\{t = (n_{a})-[r_{ab}]-(n_{b}) | t \in Graph \right\}$ is a set of triples of nodes and relations, where $n_{a}$ represents the node $a$ and $r_{ab}$ is the relation between node $a$ and node $b$. If the number of paths 
between the two target nodes is less than $k$, we add empty paths to the set until the number meets the requirement.

For each path $P_i$, we extract all the relations $r_{iab}$ and use one-hot encoder to encode them as vectors $\vec{r}_{iab}$. We than define the graph embedding between the two target nodes as:
\begin{equation}
    \vec{h}_{ij} =  \left[\sum \vec{r}_{1} ,\sum \vec{r}_{2},..., \sum \vec{r}_{k} \right]
\end{equation}

For each empty path, we use an all -1 vector to represent the summation result of its relation vector.

\subsubsection{Graph Neural Network}
%图信息embedding:我们使用了GAT（Graph Attention Network）模型来对底层知识图谱中的每个节点进行计算其embedding。GAT模型使用了masked attention机制解决了图神经网络的非谱方法（non-spectral approaches）中输入节点degree不同的问题。
%我们使用了基于link prediction的GAT模型进行节点embedding训练。我们的输入为节点以及它们的邻居节点以及他们与邻居节点的关系。对于每个节点，我们初始化了一系列vector作为每个的embedding。对于两个节点之间的关系，我们使用one-hot的embedding。首先我们会通过一个linear层来编码邻居节点本身的embedding以及关系的embedding作为feature
\begin{figure*}[htbp]
\centerline{\includegraphics[width=6.5 in]{{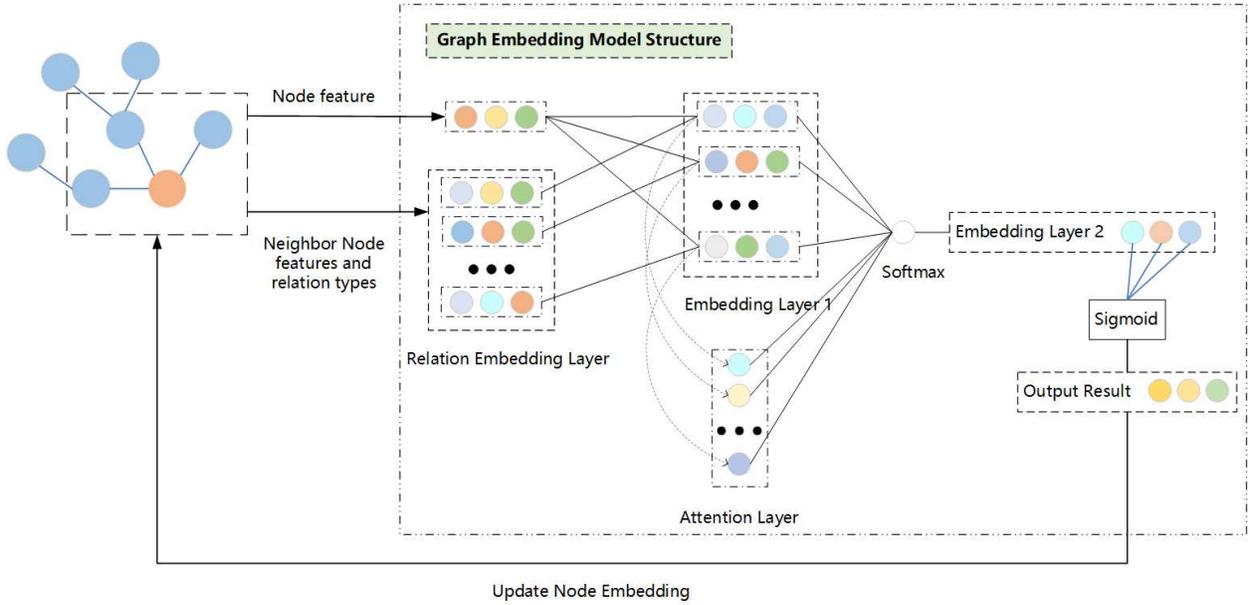}}}
\caption{The Graph embedding model structure. The input of the model is the embedding of a node, the embeddings of its neighbor nodes and the one-hot vectors representing the relation types. After completing the workflow, the model will update the new node embedding to the original graph.}
\label{gat}
\end{figure*}

In this approach, we use the Graph Neural Network based on self-attention to calculate the embedding of each entity node in the main knowledge graph. We use the masked attention mechanism which designed by Veli\v{c}kovi\'c \emph{et al.} in GAT~\cite{2018graph} to solve the  lack of flexibility for the non-spectral graph convolution networks approaches. Figure~\ref{gat} shows the model structure of the graph self-attention model.

We randomly initialize $\textbf{h} = \{\vec{h}_1, \vec{h}_2,...,\vec{h}_N\} , \vec{h}_i \in R^F $ as a set of node feature vectors, where $N$ is the number of nodes, and $F$ is a constant. We use one-hot encoder for relations. The relation between node $i$ and node $j$ are defined as $r_{ij}, r_{ij} \in R^S$, and $S$ is the number of relation categories. 

We use the graph self-attention model based on link prediction to train the embedding of nodes. 
Our input is the embeddings $\vec{h}_i$ of the target node $i$ and a set of tuples where each tuple represents one of $i$'s neighbor nodes and the its relation to $i$ denoted as $\{(\vec{h}_j, \vec{r}_{ij}) | j \in N(i)\}$, where $N(i)$ represents $i$'s neighbor nodes. 
The relation embedding layer is used to encode the embedding of the neighbor node itself and the one-hot vector of the relation into one  feature vector. It works as:
\begin{equation}
\vec{h}_{j}' = W_a \left[ \vec{h}_j,\vec{r}_{ij} \right]
\end{equation}
where $W_a \in R^{F \times (F+S)}$. Another linear transformation layer, which is parameterized by a weight matrix $W_b \in R^{F \times 2F}$, is then used to combine the embedding of $i$ with the previous layer outputs.
\begin{equation}
\vec{h}_{j}'' = \sigma \left( W_b \left[ \vec{h_i},\vec{h}_j' \right] \right)
\end{equation}

We then use a shared attention mechanism to compute the attention coefficients $e_{ij}$ for each neighbor node $j$ to measure the importance of node $j$ f node $i$:
\begin{equation}
e_{ij} = \vec{a}\vec{h}_{j}'' 
\end{equation}
where $\vec{a} \in R^{1 \times F} $ is a single-layer feed-forward neural network. We use the softmax function to normalize the coefficients as:

\begin{equation}
    \alpha_{ij} = \frac{ \exp \left(e_{ij} \right) }{ \sum_{ k \in N(i) } \exp \left(e_{ij} \right) }
\end{equation}

We compute the linear combination of the features of each neighbor node by the attention coefficients as the result of embedding. To avoid the problem of covariate shift, we apply a layer normalization function.
\begin{equation}
    \vec{h}_{i}' = \sigma \left[ W_c \mbox{LayerNorm}\left(\sum_{j \in N(i)}\alpha_{ij}[\vec{h}_{i},\vec{h}_{j}''] \right) \right]
\end{equation}
where $W_c \in R^{F \times 2F}$.

%为了解决样本比例不平衡的情况，我们使用了focal loss来进行训练
To train the model, we use a simple linear classifier to predict the relation between two nodes. To mitigate the problem of imbalanced data, we use the focal loss~\cite{lin2020focal}:
\begin{equation}
    FL(p_t) = -\alpha_t(1-p_t)^{\lambda}\log(p_t)
\end{equation}
where $p_t$ is the predicted probability of the model for category $t$, $ \lambda $ is the modulating factor and $ \alpha_t $ is the weighting factor of category $t$. We calculate the factor $\alpha_t$ by the equation:
\begin{equation}
    \alpha_t = \frac{\exp \left( n_{sum}/n_t\right)}{\sum_{s \in c} \exp\left(n_{sum}/n_s\right)}
\end{equation}
where $n_t$ is the number of occurrences for relation $t$ in the graph, and $n_{sum} = \sum n_t$ represents the total number of relations in the graph.

\subsection{DEE model based on knowledge graph}
%这一部分我们来介绍一下事件抽取的算法模型。整个模型分为X个部分：命名实体识别，实体以及文章信息embedding，图信息embedding，事件类型分类以及事件路径构建
In this section, we describe the event extraction model. The whole model is divided into three parts: named entity recognition, entity and sentence information embedding, and event path construction.

\subsubsection{Named Entity Recognition}
%命名实体识别：在命名实体识别的部分，我们使用了Transformer+CRF的方式在句子级别进行这项任务。我们将文章以句子的形式输入模型中。其中句子是由一系列word embedding组成的。
We use a Transformer+CRF~\cite{vaswani2017attention} model in the NER part. We input the document into the model in the form of a sequence of sentences $\left[s_1;s_2;,..;s_{N_{s}}\right]$. The sentence $s_i$ is represented by a sequence of token embeddings $[w_{i,1},w_{i,2},...,w_{i,{N_{w}}}]$, where $w_{i,j} \in R^{d_w}$ corresponds to the $j$th token in $i$th sentence. The hyperparameters $N_{s}$ and $N_{w}$ are the maximum lengths of the sentence sequence and the token sequence respectively, and $d_w$ is the embedding size of tokens.

%最后我们将向量通过一个CRF层输出最终的结果，并适用viterbi解码获得最好的label
After that, the Transformer layer embeds the input into a vector $h_i \in R^{d_w \times N_w}$. Finally, we us a conditional random field (CRF) layer to get the output, and apply Viterbi decoding on the output the to get the NER result.

\subsubsection{Entity Embedding}
%实体信息embedding：因为信息分散在全文中，为了获得尽可能多的信息，我们需要对实体进行全文级别的编码。我们从实体所处的位置，实体的标签以及全文的信息这三个不同的维度来对不同实体进行编码。
Because the information of the document is sparse, we encode entities at document-level to obtain as much information as possible. We encode different entities from three dimensions: the label of entities, sentence, and document.

%我们把实体和句子输入一个max-pooling层进行降维。在这个过程中,一个句子或一个实体被视为token组成的list。我们应用max-pooling的方法在每一个实体上来获得一个embedding。
We input entities and sentences into a max-pooling layer for dimensionality reduction. In this stage, a sentence or an entity is regarded as a list of tokens. We apply the max-pooling method to each entity $e_l$ and sentence $s_l$ to get the new embeddings $e_l',s_l' \in R^{d_w}$. 

%我们对每个实体进行了label type的embedding。我们的NER的输出结果包含着不同类别的信息。
We also encode the label type for each entity. The output of our NER contains different label types such as date, company name, and company id. The encoding process is
\begin{equation}
    e_i' = \mbox{LayerNorm} \left( e_i + W_l \left[ l_i \right] \right)
\end{equation}
where $l_i \in R^{N_l}$ represents the one-hot vector of the entity's label type, $N_l$ is the total number of label type categories and $W_l \in R^{d_w \times N_l}$ is a weighted matrix.

%最后，为了获得篇章级别的embedding。我们使用了一个Transformer层来编码实体与句子。最后我们获得了实体的list以及句子的list
To obtain document-level embeddings, we use a Transformer layer to encode entities and sentences. After Transformer encoding, we use the max-pooling operation to merge the entities with the same entity name into a single embedding. Finally, we get a list of entities $e^d = [e^d_1,...,e^d_{N_e}]$ and a list of sentences $s^d = [s^d_1,...,s^d_{N_s}]$, where $N_e$ is the number of distinct entities.

\subsubsection{Event Path construction}
%path expanding 

%我们将事件抽取问题转化为了一个构建entity-based directed acyclic graph的问题。对于每种类型的事件，我们都predefine了一个field list描述其中的事件元素，并以将field连接起来成为一个有向无环图。我们通过对图中的每一个节点进行path-expanding task来完成最后的事件。
We transform the event extraction problem into a problem of constructing an entity-based directed acyclic graph (EDAG) which is similar to~\cite{zheng2019doc2edag}. Compared with the simple table filling method~\cite{yang2018dcfee}, the method of constructing a graph can be divided into some path-expanding task, which is easier to handle. At the same time, the method of constructing a graph is more suitable for incorporating graph embedding information. 
For each event type, we define a field list to describe the event roles, and connect the fields in order to form a directed acyclic graph. This method allows us to extract the target event by path-expanding on each node in the graph.

%首先，我们使用max pooling对全文的句子进行编码，得到一个向量。并通过一个线性分类器来判断这篇文档的事件类型
In this part, we first use max-pooling to encode the sentences of the document to get a vector $t \in R^{d_w}$ and use a linear classifier to get the event type of this document.

%接下来，我们根据判断出的事件类型来构建出一个初始化的EDAG节点。然后我们根据预先定义的事件元素顺序对这个节点进行path-expanding 任务。在每个path-expanding task中，为了记忆之前路径中的实体，我们构建了一个memory向量。在扩展当前path后，我们会将当前实体append到memory向量。如果没有实体被选择，我们则会使用一个零向量来更新m。
Next, we construct an initialized EDAG node according to the event type. Then we expand the path on this node based on the predefined sequence of event role. In order to remember the entities in the previous path, we construct a memory vector $m$. After expanding the current path, we append the current entity to the memory vector. If no entity is selected, we use a zero-padding vector to update $m$.

%我们在path-expanding部分加入了知识图谱的embedding信息。我们使用一个列表记录在先前path-expanding任务中expand的所有实体的graph embedding，如果graph中没有这个实体，则使用一个zero-padding vecotr来表示。对每个备选实体，我们将它的graph embedding与
As shown in Figure~\ref{ee model}, we add the knowledge graph embedding information in the path-expanding task. We use a list $e^p $ to record the graph embedding of all entities expanded in the previous path-expanding task. If there is no such entity in our main knowledge graph, a zero-padding vector is used to represent it. For each candidate entity, we input its graph embedding and the list $e^p$ to an encoder to obtain a vector $g \in R^{F}$. We use the Transformer 
model with max-pooling as the encoder. We also implement a single linear layer with max-pooling approach for comparisons.
%We use two different models as the encoder, a single linear layer and a transformer model. After that, we operate the max-pooling function on the model output to get the final vector $v$.
%我们用了Transformer+max-pooling模型来作为encoder。我们也实现了一个single linear layer with max-pooling作为encoder来进行对比实验。

%对每个path-expanding任务，我们合并所有实体和memory向量，然后利用一个transformer模型来编码它们。最后，我们通过一个线性分类器来确定哪个实体会被加入成为当前path下的一个节点。
For each path-expanding task, we concatenate all the entities $e^d$ , memory vector $m$ and the graph embedding vector $v$, and feed them to a Transformer model. Finally, we use a linear classifier to determine which entity will be added as a node under the current path. If multiple entities are added to the graph in the same path-expanding task, the graph generates multiple branches at the current node.

%KG data distribution
\begin{table}[tbp]
\caption{Main Knowledge Graph data distribution}
\begin{center}
\begin{tabular}{|c|c|c|}
\hline
relation& Quantity &  Ratio\\
\hline
Branch &  98,428 & 22.68\% \\
Creditor & 1,014 & 0.23\% \\
Share Holder& 138,853 & 32.00\% \\
Invest & 4,873 & 1.12\% \\
Legal Person & 95,745 &  22.07\% \\
Pledge &  61,242 &  14.11\% \\
Managing member & 33,735 & 7.78\% \\
\hline
\end{tabular}
\label{kgdata}
\end{center}
\end{table}

%数据质量table
\begin{table}[tbp]
\caption{Labeled Data Quality}
\begin{center}
\begin{tabular}{|c|c|c|c|}
\hline
Method& 
Precision& 
Recall&
F1-score\\
\hline
 Template Matching & 
76.36\% & 
64.70\% &
70.04\%\\
 Distant Supervision & 
\textbf{92.86\%} & 
\textbf{85.71\%} &
\textbf{89.14\%}\\
\hline
  \end{tabular}
\label{dataanno}
\end{center}
\end{table}

%实验dataset分布
\begin{table}[tbp]
\caption{DEE dataset statistics and split.}
\begin{center}
\begin{tabular}{|c|c|c|c|c|}
\hline
Event Type&  Train& Val. &Test & Total\\
\hline
Equity Freeze &  407 &  183 & 106 &  696 \\
Equity Repurchase & 2,331 & 547  & 534 &  3,412 \\
Equity Overweight & 3,292 & 240 & 350 & 3,882 \\
Equity Underweight & 3,278 & 297 & 217 & 3,774 \\
Equity Pledge & 7,657 & 908 & 913 & 9,478 \\
Lawsuit &  498 & 64 & 59 & 621 \\
Total & 17,463 & 2,179 & 2,179 & 21,871 \\
\hline
\end{tabular}
\label{tabdataset}
\end{center}
\end{table}

\section{Experiments}\label{experiments}
%我们详细的描述了我们使用的数据，包括实验数据集以及知识图谱数据。我们展示了实验的结果，并且将我们的模型与state-of-the-art的模型的实验结果进行对比，包括了对总体的评价以及对single event以及multi-event分开对比。我们仔细地分析了实验结果，讨论了我们框架在当前任务上的优缺点。
In this section, we first describe the data used in our framework, including main knowledge graph data and the labeled financial announcement dataset. We then compare the experiment results from our models with these from a state-of-the-art model and further divide the test data into single event data and multi-event data to examine the differences in performance. We also carefully discuss the advantages and disadvantages of our framework.

%知识图谱
\subsection{Knowledge Graph}
We use business information of 3,000 listed companies from China's National Enterprise Credit Information Publicity System\footnote{http://www.gsxt.gov.cn/}, and construct a directed knowledge graph with 347,570 entities (companies or persons) and 441,127 relations (7 in total, such as Share Holder and Branch, mainly formed by commercial activities). In our framework, we focus on the relations among entities. The details about the relations are listed in Table~\ref{kgdata}.
%\footnote{\url{http://www.gsxt.gov.cn/}}

%%%%%%experiment reuslt 1%%%%%%%
\begin{table*}[htbp]
\caption{Event extraction results (\%).}
\label{table}
\begin{tabular}{|c|c|c|c|c|c|c|c|}
\hline
Model& EF & ER &  EO & EU & EP & LA & Total\\
& 
P.\quad\;\;  R.\quad\;\; F1& 
P.\quad\;\;  R.\quad\;\; F1&
P.\quad\;\;  R.\quad\;\; F1&
P.\quad\;\;  R.\quad\;\; F1&
P.\quad\;\;  R.\quad\;\; F1&
P.\quad\;\;  R.\quad\;\; F1&
P.\quad\;\;  R.\quad\;\; F1\\
\hline
Doc2EDAG & 
78.4\; 70.2\; 74.1 &
81.3\; 80.9\; 81.6 &
84.0\; 73.2\; \textbf{78.2} &
77.3\; 64.8\; 70.5 &
77.5\; 73.0\; 75.2 &
65.9\; 59.3\; 62.4&
79.0\; 73.9\; 76.4\\

One-hot+Tansformer & 
82.2\; 76.2\; 79.1 &
86.4\; 87.4\; \textbf{86.9} &
84.5\; 70.5\; 76.9 &
81.7\; 76.7\; 79.1 &
83.9\; 76.3\; 79.9 &
59.0\; 66.3\; 62.4 &
79.6\; 75.6\; 77.4 \\

GNN+Linear & 
82.5\; 70.2\; 75.8 &
83.4\; 79.8\; 81.6 &
84.0\; 71.8\; 77.4 &
71.6\; 64.1\; 67.7 &
74.4\; 76.7\; 75.5 &
55.4\; 56.6\; 56.0 &
77.2\; 75.2\; 76.2\\

GNN+Transformer & 
85.9\; 82.3\; \textbf{84.1} &
86.1\; 87.4\; 86.7 &
83.9\; 69.5\; 76.0 &
85.0\; 76.5\; \textbf{80.5} &
83.9\; 77.9\; \textbf{80.8} &
67.4\; 65.6\; \textbf{66.5} &
\textbf{84.4}\; \textbf{79.1}\; \textbf{81.7} \\
\hline
\end{tabular}
\label{tabexpresult}
\end{table*}

%%%%%%experiment reuslt 2%%%%%%%
\begin{table*}[tbp]
\caption{Single Event (S.) VS Multi-Event (M.) results in F1-scores (\%)}
\begin{center}
\begin{tabular}{|c|c|c|c|c|c|c|c|}
\hline
Model& EF & ER & EO & EU & EP & LA & Total\\
& 
S.\quad\;\;  M.& 
S.\quad\;\;  M.& 
S.\quad\;\;  M.& 
S.\quad\;\;  M.& 
S.\quad\;\;  M.& 
S.\quad\;\;  M.& 
S.\quad\;\;  M.\\
\hline
Doc2EDAG & 
74.9\; 64.0 &
87.7\; 62.7 &
80.1\; 74.4 &
76.5\; 60.9 &
83.5\; 69.0 &
67.6\; \textbf{58.3}&
78.4\; 64.9\\

GNN+Transformer & 
\textbf{88.1}\; \textbf{81.9} &
\textbf{92.1}\; \textbf{74.7} &
\textbf{85.3}\; \textbf{76.3} &
\textbf{79.8}\; \textbf{70.0} &
\textbf{87.8}\; \textbf{76.4} &
\textbf{69.6}\; 55.0&
\textbf{87.8}\; \textbf{76.2}\\
\hline
\end{tabular}
\label{svmdataset}
\end{center}
\end{table*}

%讲怎么用前面的方法标的数据
\subsection{Data Labelling}\label{data label exp}
In addition to the dataset in~\cite{zheng2019doc2edag}, we use the distant supervision approach described in Section~\ref{ds labelling method} to obtain 2,528 labelled lawsuit type financial announcements. To be specific, we start with a total of 14,052 lawsuit type financial announcements from Shanghai Stock Exchange and Shenzhen Stock Exchange in over 10 years (2010-2020). After applying the template matching method, 2,937 announcements contain enough event information to be used in our experiments and they include 3,197 events in total. From these announcements, we construct a basic lawsuit knowledge graph with 9,839 nodes and 12,887 relations that serves as the knowledge base for the DS method mentioned in Section~\ref{ds labelling method}, which yields the aforementioned 2,528 announcements.

Table~\ref{dataanno} shows the results of template matching and the DS labelling experiment. We randomly select 100 documents and manually check the documents to test the quality of our dataset. It suggests that the DS method improves template matching and achieves a satisfying F1-score of almost 90\%. It is worth mentioning that our released dataset based on this has been further manually checked and corrected to facilitate relevant research.

\subsection{Dataset Construction}
We keep the announcements from the Doc2EDAG dataset~\cite{zheng2019doc2edag} and the lawsuit dataset (Section~\ref{data label exp}) which include at least one entity that appears in the main knowledge graph such that the knowledge graph can potentially help. As a result, we get a new dataset with 21,871 announcements, covering 6 event types: equity freeze (EF), equity repurchase (ER), equity overweight (EO), equity underweight (EU), equity pledge (EP) and lawsuit (LA). We split the data into training, validation, and test sets following a ratio of 8:1:1 and use this dataset for all of our DEE experiments and evaluations. Details of this dataset can be found in Table~\ref{tabdataset}.

\subsection{DEE Experiment}
%我们使用了3种不同的方式引入知识图谱先验信息的embedding，并分别进行了实验。
%我们的baseline是Doc2EDAG原模型。我们在三种不同的图信息引入方式One-hot+Transformer,GAT+Linear以及GAT+Transfromer上进行了实验。GAT和One-hot是先前提到的两种不同的知识图谱先验信息编码方式。Linear和Transformer分别代表使用single Linear层和tansfromer处理输入的图信息。
In this part, we compare our models with the state-of-the-art Doc2EDAG model. Our models include the main GNN+Transformer model as well as the One-hot+Transformer and GNN+Linear models, all elaborated in Section~\ref{methodology}. GNN embedding and one-hot encoder are compared as two different methods of incorporating priori information from external knowledge graphs. With regard to embedding merging, Single-Linear layer and Transformer are evaluated.

\subsubsection{Main result}
%我们可以发现GAT+Transformer的模型获得了最好的F1-score
%在所有类别中，加入了先验信息的模型都超过了原Doc2EDAG模型。
%尤其在EF，LA两个任务中，GAT+Transformer的模型F1-score超过了Doc2EDAG模型5.3%，6.3%
As we can see in Table~\ref{tabexpresult}, the overall performance of our models that incorporate knowledge graph information beats Doc2EDAG, with the GNN+Linear model as the only exception that has a slightly lower F1-score and a higher recall. This indicates that proper integration of relational knowledge can improve the performance of the event extraction task.

It is worth noting that GNN+Transformer, as the best-performing model with an overall F1-score of 81.7\%, surpasses Doc2EDAG by a relatively large margin (5.4\% for precision, 5.2\% for recall, and 5.3\% for F1-score). In particular, for the EF, ER, EU, and EP categories, the GNN+Transformer model demonstrates significant improvements by 10.0\%, 6.1\%, 10.0\%, and 5.6\%, respectively. This may be related with the characteristics of the events in these categories -- entities within a document are often related intrinsically. For instance, in the case of equity pledge, the pledgee is more likely to be a bank that already has pledge relations with many companies, which is recorded in our relational knowledge graph. Another example is equity overweight which happens when a listed company's major stake-holders plan to purchase more of its shares and these stake-holders are usually captured in the knowledge graph.

When compared with the other two graph based models, the GNN+Transformer model also shows improvements in F1-scores. Its 4.3\% improvement over One-hot+Transformer implies that GNN is more effective in capturing graph information, as apposed to the intuitive feature engineering over the graph, while the 5.5\% improvement over GNN+Linear suggests that the Transformer model with its self-attention mechanism better encodes the information, compared with the simple linear layer.

%特别的，在XXX,XXX以及XXX类别上，GNN+Transformer模型对比baseline有明显的提升. 
%对比另外两种实现方法，我们的模型在整体的F1-score上有着更好的表现。

%由于我们抽取了知识图谱中的关系信息，所以我们的模型能更有效的在存在更多关系信息的事件类型中起作用，例如EF,ER以及EP. 

\subsubsection{Single Event vs Multi-Event}
We examine the performance of GNN+Transformer and Doc2EDAG on single event documents (each containing only one event) and multi-event documents (each containing more than one event), by further splitting the test data and calculating the corresponding F1-scores at document level (shown in Table~\ref{svmdataset}). Overall, our model improves Doc2EDAG by 9.4\% and 11.3\% in the single event dataset and the multi-event dataset, respectively. For the multi-event documents, the improvement is higher. A possible reason is that these documents with multiple events usually contain multiple entities and entities across events may cause confusions for the Doc2EDAG model, while this might be mitigated by our model based on a relational knowledge base.

%我们在Doc2EDAG和GNN+Transformer两个模型上进行了实验。我们可以发现，在single event数据集中，我们的模型比Baseline提升了9.4，而在multi-event中，我们的模型相比baseline有11.3%的更好表现

\subsubsection{Discussion}
%我们的模型通过引入知识图谱的关系信息，在实验中打败了state-of-the-art的baseline模型。
%同时，我们也发现了GNN+Transformer的形式能够更有效的embedding知识图谱的信息。相比于我们使用的两个节点之间路径的one-hot embedding，GNN能够隐性的学习一个节点与它所有邻居节点的关系以及节点在图中的位置。此外，one-hot embedding无法描述序列信息，而我们的GNN模型通过self-attention解决了这一问题。这些GNN的特性导致了它在任务中获得了更好的效果。
%相比其它rnn，Transformer模型具有可并行以及可解释性高的特点，可以提高模型的实用性

All of the above experiments suggest that the combination of GNN and Transformer can more effectively embed knowledge graph information, comparing with the other two combinations. Different from the plain one-hot encoding of the path between two nodes, GNN can implicitly learn the relation between a node and all its neighbor nodes, as well as the position of the node in the graph. In addition, one-hot encoder cannot describe sequence information, but our GNN model over come this problem through self-attention. Transformer, as a seq2seq model, can encode the relations between two nodes well when there are multiple nodes in between and the paths are sequential. Besides, compared with other Recurrent Neural Networks, the Transformer model can be easily paralleled and offers higher interpretability, making it more applicable in solving real-world problems~\cite{vaswani2017attention}.

\section{Conclusion and Future Work}\label{conclusion and future work}
%本文提出了一个基于知识图谱的事件抽取框架。相比state-of-the-art的baseline，我们的框架获得了更好的效果。为了能让模型对现实场景更有意义，我们在原数据集的基础上构建了一个诉讼类型事件集。在嵌入知识图谱的方法上，我们发现我们的模型GNN+Transformer的实现方式好于其它两种我们实现的方法。实验结果证明先验关系知识的引入对于事件抽取任务有效果
In this paper, we propose a knowledge graph based extraction framework. While a straightforward way of engineering the graph into features can effectively improve the state-of-the-art baseline in financial announcement extraction, our GNN+Transformer model that better integrates the graph information with limited human intervention archives a considerable 5.3\% improvement. In order to make the model more applicable in the real scenario, we construct a lawsuit type event dataset and merge it with a public dataset. While all our models that incorporate the knowledge graph outperform the state-of-the-art baseline, the GNN+Transformer model that learns more precise relational knowledge achieves the best performance.

%由于我们的框架在事件抽取任务上取不错的效果，而其它NLP tasks例如关系抽取，句子级别的事件抽取中，关系信息也可能能够帮助信息的提取。我们plan to 应用一些更general的知识图谱在更多的nlp task上。
As a future direction, we plan to apply the framework on tasks including relation extraction and sentence-level event extraction where relational prior knowledge can also be helpful. Besides, we will process documents in other domains with our framework augmented with general knowledge graphs including DBpedia and Freebase.

\section{Acknowledgments}
We thank the anonymous reviewers for their constructive suggestions. This project is sponsored by Shanghai Sail Program (No.19YF1433800) and the Key Projects of Shanghai Soft Science Research Program (No.20692194300). This article has been accepted to IEEE Big Data 2020 Comference.

\bibliographystyle{IEEEtran}
\bibliography{references}

\end{document}